\documentclass[11pt]{article}

\usepackage{graphics}

\usepackage{algorithmic}
\usepackage{algorithm}

\newcommand{\algo}[3]{
  \begin{algorithm}
    \caption{#1}
    \label{algo:#2}
    \begin{algorithmic}
      #3
    \end{algorithmic}
  \end{algorithm}
}

\def\R{\mbox{I\hspace{-.15em}R}}

\def\begeq{
\begin{equation}
}
\def\endeq{\end{equation}
}

\def\begeqa{
\begin{eqnarray}
}
\def\endeqa{\end{eqnarray}
}

\newcommand{\image}[2]{
\resizebox{#2cm}{!}{\includegraphics{./Images/#1.eps}}
}

\title{Modular self-organization}

\author{%Machine Learning, Decision Theory, Reinforcement Learning
       Bruno SCHERRER\\       
       {LORIA, BP 239}\\
       {54506 Vand\oe{}uvre-l\`{e}s-Nancy}\\
       {France} \\
       \texttt{scherrer@loria.fr}
}

\begin{document}

\maketitle

\begin{abstract}
  The aim of this paper is to provide a sound framework for addressing
  a difficult problem: the automatic construction of an autonomous
  agent's modular architecture. We combine results from two apparently
  uncorrelated domains: Autonomous planning through Markov Decision
  Processes and a General Data Clustering Approach using a kernel-like
  method.  Our fundamental idea is that the former is a good framework
  for addressing autonomy whereas the latter allows to tackle
  self-organizing problems.
% 
% Using a repartitioning algorithm derived
% from a recent theoretical analysis by \cite{munos99rates} and an
% on-line version of the Dynamic Cluster algorithm originally
%  introduced by \cite{diday73}, we demonstrate the power of such an
%  approach: 
  Indeed, we derive a modular self-organizing algorithm in which an
  autonomous agent learns to efficiently spread $n$ planning problems
  over $m$ initially blank modules with $m<n$.  
\end{abstract}

\section*{Introduction}

This paper addresses the problem of building a long-living autonomous
agent; by \emph{long-living}, we mean that this agent has a large
number of relatively complex and varying tasks to perform. Biology
suggests some ideas about the way animals deal with a variety of
tasks: brains are made of specialized and complementary areas/modules;
skills are spread over modules.  On the one hand, distributing
functions and representations has immediate advantages: parallel
processing implies reaction speed-up; a relative independence between
modules gives more robustness. Both properties might clearly increase
the agent's efficiency. On the other hand, the fact of distributing a
system raises a fundamental issue: how does the organization process
of the modules happen during the life-time ?

There has been much research about the design of modular intelligent
architectures (see for instance \cite{theocharous00learning}
\cite{hauskrechthierarchical} \cite{blanchet96}
\cite{kaelbling93hierarchical}). It is nevertheless very often the
(human) designer who decides the way modules are connected to each
other and how they behave with respect to the others.  Few works study
the construction of these modules. To our knowledge, there are no
effective works about modular self-organisation except for reactive
tasks (stimulus-response associations) \cite{jacobstask}
\cite{lange94growing} \cite{digney-emergent}.

This paper proposes an architecture in which the partition in
functional modules is automatically computed. The most significant
aspect of our work is that the number $m$ of modules is fewer than the
number $n$ of tasks to be performed. Therefore, the approach we
propose involves a high-level clustering process where the $n$ tasks
need to be ``properly'' spread over the $m$ modules.

Section 1 introduces what we consider as the theoretical foundation for
modelling autonomy: Markov Decision Processes. Section 2 presents the
state aggregation technique, which allows to tackle difficult
autonomous problems, that is large state space Markov Decision
Processes.  Section 3 describes the Kernel Clustering approach: it
will stand as a theoretical basis for addressing self-organization.
Kernel Clustering will indeed lead to a generalization of the state
aggregation technique, which we will interpret as a modular
self-organization procedure. Finally, Section 5 will present
empirical results about the self-organization of an autonomous agent
that has to navigate in a continuous environment.

\section{Modelling A Mono-Task Autonomous Agent}

Markov Decision processes \cite{Puterman94} provide the theoretical
foundations of challenging problems such as planning under uncertainty
and reinforcement learning \cite{sutton98}. They stand for a
fundamental model for sequential decision making and they have been
applied to many real worls problem \cite{sutton97significance}. This
section describes this formalism and presents a general scheme for
approaching difficult problems (that is problems in large domains).

\subsection{Markov Decision Processes}

%APPLICATIONS ???
%In this section by describing the Markov Decision Process formalism.
%We then describe an approximation scheme for difficult (large domain) planning problems.

%\subsection{Markov Decision Processes}

A Markov Decision Process (MDP) is a controlled stochastic process satisfying the Markov
property with rewards (numerical values) assigned to state-control
pairs\footnote{Though our definition of reward is a bit restrictive
  (rewards are sometimes assigned to state transitions), it is not a
  limitation: these two  definitions are equivalent.}. Formally,
an MDP is a four-tuple $\langle S, A, T, R \rangle$ where $S$ is the
\emph{state space}, $A$ is the \emph{action space}, $T$ is the
\emph{transition function} and $R$ is the \emph{reward function}.  
%We assume that $S$ and $A$ are finite. 
$T$ is the state-transition
probability distribution conditioned by the control\,:
\begeq
T(s,a,s')\stackrel{\small{def}}{=}  \mbox{Pr}(s_{t+1}=s'|s_t=s,a_t=a) \nonumber
\endeq
$R(s,a) \in \R$ is the instantaneous reward for taking action $a \in A$ in state
$S$.

The usual \emph{MDP problem} consists in finding an \emph{optimal policy}, that
is a mapping $\pi:S \rightarrow A$ from states to actions, that
maximises the following performance criterion, also called value
function of policy $\pi$:
\begeq
V^{\pi}(s)=\mbox{E}\left[ \sum_{t=0}^{\infty} \gamma^t.R(s_t,\pi(s_t)) | s_0=s \right]
\endeq
It is shown \cite{Puterman94} that there exists a unique optimal
value function $V^*$ which is the fixed point of the following
contraction mapping 
$B^*$ (called Bellman operator):
\begeq
\left[B^*.f\right](s) = \max_a \left( R(s,a) + \gamma.\sum_{s'} T(s,a,s').f(s') \right) 
\label{BellmanOp}
\endeq
Once an optimal value function $V^*$ is computed, an optimal
policy can immediately be derived as follows:
\begeq
\pi^*(s)=\mbox{arg}\max_a \left( R(s,a) + \gamma.\sum_{s'} T(s,a,s').V^*(s')\right)
\endeq
Therefore, solving an MDP problem amounts to computing the optimal
value function.  Well-known algorithms for doing so are \emph{Value
  Iteration} and \emph{Policy Iteration} (see \cite{Puterman94}).
Their temporal complexity dramatically grows with the number of states
\cite{littman95}, so they can only be applied to relatively simple
problems.
%Large state space MDPs thus requires the use
%of approximation.

%\subsection{Approximations in Realistic problems}

\subsection{Addressing a Large State Space MDP}

In very large domains, it is impossible to solve an MDP exactly, so we
usually address a complexity/quality compromise. Ideally, an
approximate scheme for MDPs should consist of a set of
\emph{tractable} algorithms for
\begin{itemize}
\item computing an approximate optimal value function
\item evaluating (an upper bound of) the approximation error
\item improving the quality of approximation (by reducing the approximation error) while constraining the complexity.
\end{itemize}
The first two points are the fundamental theoretical bases for sound
approximation. The third one is often interpreted as a learning
process and corresponds to what most Machine Learning researchers
study. For convenience, we respectively call these three procedures
$Approximate()$, $Error()$ and $Learn()$. 
The use of such an approximate scheme is sketched by algorithm \ref{algo:ml}:
\algo{A general approximation scheme for a large state MDP}{ml}{

 \REQUIRE a large state space MDP $\cal M$ and an initial approximation $\widehat{\cal M}$.
 \ENSURE a good approximate value function $\widehat{V}^*$.

\STATE
  \WHILE {$Error({\widehat{\cal M}},\cal M)$ goes on diminishing}
   \STATE  ${\widehat{\cal M}} \leftarrow Learn(\widehat{{\cal M}},\cal M)$
  \ENDWHILE
  \STATE $\widehat{V}^* \leftarrow Approximate(\widehat{{\cal M}})$
}
One successively applies the $Learn()$ procedure in order to
minimize the approximation error; when this is done, one can
compute a good approximate value function.
Next section describes an example of such a set of
procedures for practically approximating a large state space MDP.

\section{The State Aggregation Approximation}

\label{SA}

This section reviews an example of approximation scheme for solving
large state space MDPs.  The class of approximations we consider is
the state aggregation approximation, that is approximate models
in which whole sets of states are treated as if they had the same
parameters and underlying values.

Given an MDP ${\cal M}=\langle S,A,T,R \rangle$, the state aggregation
approximation formally consists in introducing the MDP ${\widehat{\cal
    M}}=\langle \widehat{S},A,\widehat{T},\widehat{R} \rangle$ where
the state space $\widehat{S}$ is a partition of the real state space
$S$. Every element of $\widehat{S}$, which we call macro-state, is a
subset of $S$ and every element of $S$ belongs to one and only one
macro-state.  Conversely, every object defined on $\widehat{S}$ can
be seen as an object of $S$ which is constant on every macro-state.
The number of elements of $\widehat{S}$ can be chosen little enough so
that it is feasible to compute the approximate value function of the
approximation ${\widehat{\cal M}}$.

Using some recent results by \cite{munos99rates}, we are going to
describe how the procedures $Approximate()$, $Error()$ and $Learn()$
(introduced in previous section) can be defined.

\subsection{Computing an Approximate Solution}

When doing a state aggregation approximation, natural choices
%\footnote{We want the interpolation error to be as
%  little as possible and we need to verify constraints such as
%  $\sum_{\widehat{s_2}} T(\widehat{s_1},a,\widehat{s_2})=1$} 
for the approximate parameters 
$\widehat{R}$ and $\widehat{T}$ are the averages of the real parameters
on each macro-state:
\begeq
\label{ParamEst}
\left\{
\begin{array}{rcl}
\widehat{R}(\widehat{s},a)& =& \frac{1}{\left|\widehat{s}\right|}. \sum_{s \in \widehat{s}} R(s,a) \\
\widehat{T}(\widehat{s_1},a,\widehat{s_2})&=&\frac{1}{\left|\widehat{s_1}\right|}. \sum_{(s,s') \in \widehat{s_1}\times\widehat{s_2}} T(s,a,s')
\end{array}
\right.
\endeq
From these,  an approximate value function $\widehat{V}^*$ can be computed: it is
the fixed point of the approximate Bellman operator $\widehat{B}^*$
(defined on $\widehat{S}$):
\begeq
\left[\widehat{B}^*.f\right](\widehat{s}) = \max_a \left( \widehat{R}(\widehat{s},a) + \gamma.\sum_{\widehat{s'}} \widehat{T}(\widehat{s},a,\widehat{s'}).f(\widehat{s'}) \right)\nonumber
\endeq

This constitutes the $Approximate()$ procedure in the state aggregation approach.

\subsection{Bounding the Approximation Error}

Let $B^*$ be the exact Bellman operator of ${\cal M}$ (see eq. \ref{BellmanOp}).
Let $V^*$ be the real value function. In practice, we
would like to evaluate how much the approximate
value function $\widehat{V}^*$ differs from the 
real value function $V^*$, i.e. we want to compute the
\emph{approximation error} on each macro-state $\widehat{s}$:
\begeq
E_{app}(\widehat{s}) \stackrel{\small{def}}{=}\max_{s \in \widehat{s}}|V^*(s)-\widehat{V}^*(s)|
\label{errapp}
\endeq
The authors of \cite{munos99rates} show that the \emph{approximation
  error} depends on a quantity they call \emph{interpolation error} which
is easier to evaluate:
\begeq
E_{int}(\widehat{s}) \stackrel{\small{def}}{=} \max_{s \in \widehat{s}} |\widehat{B}^*.V^*(s) - B^*.V^*(s)| %= \max_{s \in \widehat{s}}|\widehat{B}^*.V^*(s) - V^*(s)|
\label{errint}
\endeq
The \emph{interpolation error} is the error due to one approximate
mapping $\widehat{B}^*$ of the real value function $V^*$ ; it measures
how the approximate parameters $(\widehat{R}(\widehat{s},a),$ $\widehat{T}(\widehat{s},a,.))$
locally  differ from the real parameters $(R(s,a),T(s,a,.))$.
Indeed, it can be  shown that for some constant $K$
\begeqa
E_{int}(\widehat{s}) & \leq & \max_{a \in A,s \in \widehat{s}} \left| R(s,a) - \widehat{R}(s,a) \right| \label{bornesuperrint} \\
&  +& K.\max_{a \in A,s \in \widehat{s}} \left( \sum_{s' \in S} \left| T(s,a,s')-\widehat{T} (s,a,s')\right| \right) \nonumber
\endeqa
%In practice, it is straightforward to reduce the \emph{interpolation
%  error} (e.g. by putting more resources), but our real concern is to
%reduce the approximation error. 
%The work presented in
%\cite{munos99rates} shows the link between both errors: 
We can deduce from equations  \ref{ParamEst} and \ref{bornesuperrint} an upper bound $\overline{E_{int}}(\widehat{s_1})$ of the
\emph{interpolation error} on the macro-state $\widehat{s_1}$:
\begeq
\overline{E_{int}}(\widehat{s_1})=\overline{\Delta R} (\widehat{s_1}) + K.\sum_{\widehat{s_2} \in \widehat{S}}\overline{\Delta T}(\widehat{s_1},\widehat{s_2})
\endeq
with
$$\overline{\Delta R} (\widehat{s})=\frac{1}{\left|\widehat{s}\right|}.\max_{(s,s') \in \widehat{s}}|R(s,a)-R(s',a)|$$ 
and
$$\overline{\Delta T} (\widehat{s_1},\widehat{s_2})=\frac{1}{\left|\widehat{s_1}\right|}.\max_{(s_1,s'_1) \in \widehat{s_1}}|\sum_{s_2 \in \widehat{s_2}} T(s_1,a,s_2)-T(s_1',a,s_2)|.$$

Once we have an upper bound of the \emph{interpolation error}, the authors of \cite{munos99rates} show that an upper bound
$\overline{E_{app}}(\widehat{s})$ of $E_{app}(\widehat{s})$ is the fixed point of the
following contraction mapping:
\begeq
\left[ \widehat{E}.f \right](\widehat{s_1}) = \overline{E_{int}}(\widehat{s_1}) + \max_a \left( \gamma.\sum_{\widehat{s_2}} \widehat{T}(\widehat{s_1},a,\widehat{s_2}).f(\widehat{s_1}) \right) 
\label{E}
\endeq

We thus have an $Error()$ procedure.

\subsection{Improving the Approximation}

%The previous subsection shows how to measure the approximation error
%when one uses the state aggregation technique with a given partition
%$\widehat S$.

Finally, this subsection explains how one might improve a state
aggregation approximation by iteratively updating the partition $\widehat
S$.

The authors of \cite{munos99rates} introduce the notion of \emph{influence}
$I_{S_0}(\widehat{s})$ of the \emph{interpolation error at macro-state $\widehat{s}$} on the
\emph{approximation error} over a subset $S_0 \subset \widehat{S}$:
\begeq
I_{S_0}(\widehat{s}) \stackrel{\small{def}}{=} \frac{\partial \sum_{\widehat{s'} \in S_0} \overline{E_{app}}(\widehat{s'})}{\partial \overline{E_{int}(\widehat{s})}}
\endeq
They prove that the \emph{influence} $I_{S_0}$ is the fixed point
of the following contraction mapping:
\begeq
\left[D.f\right](\widehat{s})  = 
\left\{
\begin{array}{l}
1 \mbox{ iff }\widehat{s} \subset S_0 \\
0 \mbox{ iff }\widehat{s} \not\subset S_0
\end{array}
\right.
+  \gamma.\sum_{\widehat{s'}} \widehat{T}(\widehat{s'},\pi_{err}(\widehat{s'}),\widehat{s}).f(\widehat{s'}) 
\label{D2}
\endeq
where $\pi_{err}(\widehat{s})=\mbox{arg}\max_a \sum_{s'}
\widehat{T}(\widehat{s},a,\widehat{s'}).E_{app}(\widehat{s'})$ (see \cite{munos99rates} for more
details).

%We now briefly discuss how the notion of influence can help to improve the partition $\widehat{S}$.
Say we update the partition for some macro-state $\widehat{s}$ (e.g. we divide
$\widehat{s}$ in two new macro-states). The \emph{interpolation error}
$\Delta \overline{E_{int}}(\widehat{s})$ will change and a gradient argument shows the effect this will have on the
\emph{approximation error}:
\begeq
\Delta \left( \sum_{s' \in S_0} \overline{E_{app}}(\widehat{s'}) \right) \simeq I_{S_0}(\widehat{s}).\Delta \overline{E_{int}}(\widehat{s})
\endeq
Using this analysis, we are able to predict the effect that locally
refining (or coarsening) the partition $\widehat{S}$ has on the
approximation error we want to minimize.  This allows to efficiently
and dynamically balance resources of the approximation over the state space.

This constitutes a $learn()$ procedure for the state aggregation approximation.
Experimental demonstrations of a similar approach can be found in \cite{munos99variable}.
%suggests that the interpolation error on
%$s$ by $\Delta \overline{E_{int}}(s)$ (i.e.  augmenting the resources
%for describing a state $s$) will reduce the \emph{approximation error}
%on $S_0$ as follows:

\section{Kernel Clustering}

So far, we have recalled recent results for approximating \emph{a
  unique} large state space MDP.  When trying to model a
\emph{long-living} autonomous agent, it is more realistic to consider
that it does not only have one problem (one MDP) to solve but rather
many (if not an infinity): $({\cal M}_i)_{1 \leq i \leq n}=(\langle
S,A,T_i,R_i \rangle)_{1 \leq i \leq n}$. In order to address such a
case, we first need to present the Kernel Clustering paradigm. This is
what we do in the remaining of this section.

\subsection{Definitions}

In \cite{diday73}, the author introduces an abstract generalization of
vector quantization, which he calls Kernel Clustering.  Indeed, the author
argues that, in general, a clustering problem is based on three elements:
\begin{itemize}
\item $(x_i)_{i \in I}$: a set of data points taken from a data space $X$
\item $\{L_1,..,L_m\}$: a set of kernels taken from a kernel space $\cal L$
\item $d:X\times{\cal L} \rightarrow \R^+$: A distance measure between any data point and any kernel. The smaller the distance $d(x,L)$, the more  $L$ is   \emph{representative} of the point $x$. 
\end{itemize}
Given a set of kernels $\{L_1,..,L_m\}$, a data point $x$ is naturally associated to its most \emph{representative} kernel $L(x)$, i.e. the one
that is the closest according to distance $d$:
\begeq
L(x)=\mbox{argmin}_{L \in \{L_1,..,L_m\}} d(x,L)
\endeq
Conversely, a set of kernels $\{L_1,..,L_m\}$ naturally induces a
partition of the data set $(x_i)_{i \in I}$ into $m$ classes $\{C_1,..,C_m\}$, each class corresponding to a kernel:
\begeq
\label{kerneltopart}
\forall j \in (1,..,m), C_j=\{(x_i)_{i \in I} ; L(x_i)=L_j\} 
\endeq

Given a data space, a data set, a kernel space and a distance $d()$,
the goal of the Kernel Clustering problem is to find the set of kernels
$\{L_1^*,..,L_m^*\}$ that minimizes the distortion $D$ for the data set
$(x_i)_{i \in I}$, which is defined as follows:
\begeq
D=\sum_{i \in I} d(x_i,L(x_i))=\sum_{j=1}^m \sum_{x \in C_j} d(x,L_j)
\endeq
In other words, solving a clustering problem consists in finding the
kernels that are the most representative of the data set.

For instance, the well-known vector quantization problem is a particular case of Kernel Clustering where 
\begin{itemize}
\item the set of kernels $\cal L$ and the data space $X$ are $\R^n$
\item the distance $d(x,L)$ is the Euclidean norm $\|x-L\|$.
\end{itemize}
As we will see in the next sections, the power and the richness of the
Kernel Clustering approach over simple vector quantization comes essentially from the fact that kernels and data need not be in the same space.

\subsection{The Dynamic Cluster Algorithm}

An interesting observation about the Kernel Clustering approach is the
following fact: the Dynamic Cluster algorithm \cite{diday73} (see
algorithm \ref{algo:dc}) for (suboptimally) optimizing the set of kernels
is the exact generalization of the \emph{batch k-means} algorithm, which
(suboptimally) solves the vector quantization problem (see
\cite{forgy65} and \cite{diday73}).
\algo{The dynamic cluster algorithm}{dc}{
\REQUIRE{A data set $(x_i)_{i \in I}$}
\ENSURE{A set of kernels $\{L_1,..,L_m\}$ that optimizes the clustering (i.e. that minimizes the distortion)}
\STATE \textbf{Initialization:}
\STATE Let $\{C_1,..,C_m\}$ be any partition of the data set
\STATE \textbf{Iterations:}
\REPEAT
 \STATE \textbf{1.} Find the best set of kernels corresponding to the partition $\{C_1,..,C_m\}$:
 \FOR {$j$ from $1$ to $m$}     
  \STATE $ L_j \leftarrow \mbox{argmin}_{L \in {\cal L}} \sum_{x \in C_j} d(L,x)$
 \ENDFOR 
 \STATE \textbf{2.} Find the partition $\{C_1,..,C_m\}$ corresponding to the kernels $\{L_1,..,L_m\}$:
 \FOR {$j$ from $1$ to $m$}
  \STATE $C_j \leftarrow \{(x_i)_{i \in I} ; L(x_i)=L_j\}$
 \ENDFOR
\UNTIL{there is no more change in the partition $\{C_1,..,C_m\}$}
}
This is an iterative process which consists of two complementary steps:
\begin{itemize}
\item Given a partition $\{C_1,..,C_m\}$, find the best corresponding kernels $\{L_1,..,L_m\}$ 
\item Given a set of kernels $\{L_1,..,L_m\}$, deduce the corresponding partition $\{C_1,..,C_m\}$.
\end{itemize}
If the latter step is straightforward (one just applies equation \ref{kerneltopart}), the former is itself an optimization problem which can be very difficult. In a general purpose, it might be easier to use an \emph{on-line} version of  the Dynamic Cluster algorithms\footnote{As it is often easier to use \emph{on-line} version of the \emph{k-means} algorithm} (see algorithm \ref{algo:dcol}).
\algo{The on-line dynamic cluster algorithm}{dcol}{

 \REQUIRE{A data set $(x_i)_{i \in I}$}
 \ENSURE{A set of kernels $\{L_1,..,L_m\}$ that optimizes the clustering (i.e. that minimizes the distortion)}

\STATE \textbf{Initialization:}
\STATE Let $\{L_1,..,L_m\}$ be any set of kernels
\STATE \textbf{Iterations:}
 \WHILE{the distortion goes on diminishing} 
\STATE Randomly pick a data point $x$ from the data set
\STATE Find the kernel the most representative kernel of $x$:
$$ L \leftarrow L(x)=\mbox{argmin}_{L' \in \{L_1,..,L_m\}} d(x,L') $$
\STATE Update $L$ so that $d(x,L)$ diminishes
\ENDWHILE
}
The resulting algorithm becomes simple and intuitive: for each piece of data $x$, one finds its most representative kernel $L$, and one updates $L$ so that
it gets even more representative of $x$. Little by little, one might expect
that such a procedure will minimize the global distortion and eventually give 
a good clustering.

\section{Modular Self-Organization For a Multi-Task Autonomous Agent}

This section is going to show how the (apparently uncorrelated) Kernel
Clustering paradigm can be used to formalize a modular
self-organization problem in the MDP framework, the algorithmic
solution of which will be given by the on-line Dynamic Cluster
procedure (algorithm \ref{algo:dcol}).

%Consider an MDP.
% So far, we have only concentrated on a unique
%planning task. T
If one carefully compares the general learning scheme we have
described in order to address a large state space MDP (algorithm
\ref{algo:ml}) and the on-line Dynamic Cluster procedure (algorithm
\ref{algo:dcol}), one can see that the former is a specific case of
the latter. More precisely, algorithm \ref{algo:ml} solves a simple
Kernel Clustering problem where
\begin{itemize}
\item the data space is the space of all possible MDPs and the data set is a unique task corresponding to an MDP $\cal M$
\item the kernel space is the space of all possible approximations and there is one and only one kernel:  $\widehat{\cal M}$
\item the distance $d$ is the $Error()$ function.
\end{itemize}
This observation suggests to make the following parallel between
Kernel Clustering and MDPs:
\begin{center}\begin{tabular}{|c|c|}
\hline
Kernel Clustering & Markov Decision Processes\\
\hline
\hline
Data space & Space of all possible MDPs \\
Data set & A set of tasks \\
Kernel space & Space of approximate models \\
Kernel & An approximate model \\
distance & Approximation error \\
%A  representative kernel & A Approximate model \\
\hline
\end{tabular}
\end{center}
The transpositon of the on-line Dynamic Cluster into the MDP
framework (algorithm \ref{algo:so}) therefore allows us to tackle
a difficult problem:
\algo{Modular Self-Organization}{so}{

 \REQUIRE{A set of MDPs $({\cal M}_i)_{i \in I}$}
 \ENSURE{A set of approximate models $(\widehat{\cal M}_1,..,\widehat{\cal M}_m)$ that globally minimizes the approximation error }

\STATE \textbf{Initialization:}
\STATE Let $(\widehat{\cal M}_1,..,\widehat{\cal M}_m)$ be any set of approximate models
\STATE \textbf{Iterations:}
 \WHILE{the global approximation error goes on diminishing} 
\STATE Randomly pick a task  $\cal M$ from the set of MDPs
\STATE Find the best module for solving $\cal M$:
$$ \widehat{\cal M} \leftarrow \mbox{argmin}_{\widehat{\cal M}' \in \{\widehat{\cal M}_1,..,\widehat{\cal M}_m\}} Error(\widehat{\cal M}',{\cal M}) $$
\STATE ${\widehat{\cal M}} \leftarrow Learn(\widehat{{\cal M}},{\cal M})$
\ENDWHILE

}
Finding a small set of approximate models that globally minimize the
approximation error for a large set of MDPs.
The result of such an approach can really be seen as a modular architecture.
Indeed, every time a task (even a new task) is given to such a system,
all kernels/modules can compute their approximation error and the best
module for solving the task is the module that makes the minimal
error.

\section{An Experiment of Modular Self-Organization}

\begin{figure}[htbp]
\centerline{\image{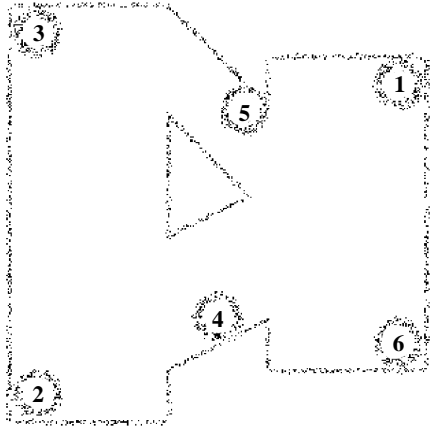}{6}}
\label{env}
\caption{A Multi-Task Environment with six goal zones}
\end{figure}

This final section provides an illustration of the Modular
Self-Organization algorithm \ref{algo:so} where the number of tasks
$n$ equals $6$ and the number of modules $m$ is $3$.  We illustrate
our approach on a navigation problem\footnote{Our self-organization
  algorithm is not limited to a navigation context; it can
  theoretically be applied to any problem which can be formulated in
  the MDP framework}. An agent has to find its way in a continuous
environment. This environment consists in 2 rooms and 2 corridors (see
figure \ref{env}). The set of states is the continuous set of
positions $(x,y) \in (0;10)^2$ in the environment.  The actions are
the 8 cardinal moves (amplitude $0.1$), whose effects is slightly
corrupted with noise (amplitude $0.03$ and random direction). Six
areas, denoted as circles in figure \ref{env} are possible goals. One
notifies an agent it has reached a goal by giving him a strict
positive reward ($+1$). One also gives a negative reinforcement ($-1$)
when the agent hits a wall. All the other situations have a zero
reward.  Note that when an agent acts optimally in such a task, it
only receives a reward when it reaches the goal.

\begin{figure}[htbp]
\centerline{\image{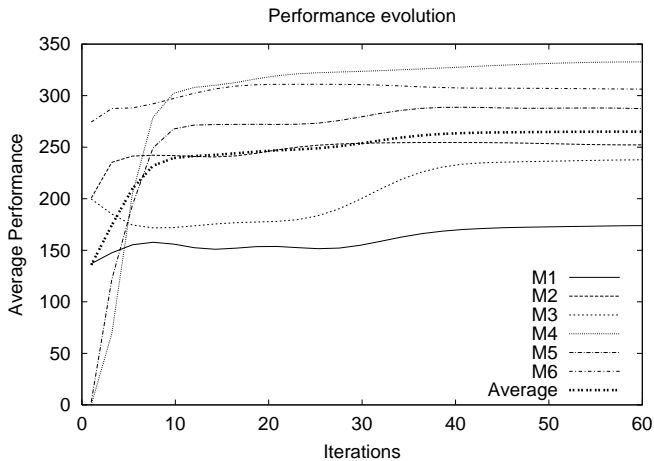}{9}}
\caption{Performance evolution during the Modular Self-Organization algorithm: for each of the six MDPs (and for the average cumulative
rewards for all six), we see that the system performance is monotonically increasing.}
\label{perfs}
\end{figure}

We use the six goal areas in order to define six MDPs/tasks.  Each of
these tasks involves going from one zone to another. The following
table sums them up:
%\vspace{.1cm}
\begin{center}\begin{tabular}{|c||c|c|}
\hline
MDP & Start & Goal\\
\hline
\hline
${\cal M}_1$  &  zone $2$ &  zone $1$ \\
${\cal M}_2$  &  zone $3$ &  zone $2$ \\
${\cal M}_3$  &  zone $4$ &  zone $3$ \\
${\cal M}_4$  &  zone $5$ &  zone $4$ \\
${\cal M}_5$  &  zone $6$ &  zone $5$ \\
${\cal M}_6$  &  zone $1$ &  zone $6$ \\
\hline
\end{tabular}
\end{center}
%\vspace{.1cm}
We have applied the Modular Self-Organization procedure (algorithm
\ref{algo:so}) with $3$ kernels/modules, and with the $Error()$ and
$Learn()$ functions described in section \ref{SA}.  Figure \ref{perfs}
shows that the performances (obtained with $500$ simulated runs for
each single task) of the system grow for the six tasks.  
\begin{figure}[htbp]
\centerline{\image{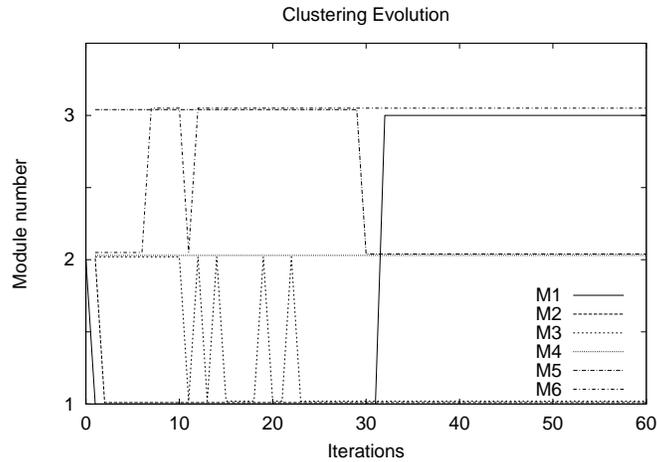}{9}}
\caption{Evolution of the clustering process: at each iteration, each task is naturally
associated to one of the six modules (the one that makes the minimal error); this spread over eventually stabilizes: at the end, module $1$ deals with ${\cal M}_2$ and  ${\cal M}_3$, module $2$ deals with ${\cal M}_4$ and ${\cal M}_5$, and module 3 deals with ${\cal M}_1$ and ${\cal M}_6$.}
\label{classif}
\end{figure}
Figure
\ref{classif} shows that the clustering (i.e. the spreading of
expertise over the $3$ modules) eventually stabilizes to an
interesting clustering: each of the eventual modules deals with two
tasks.
\begin{figure}[htbp]
\centerline{
\image{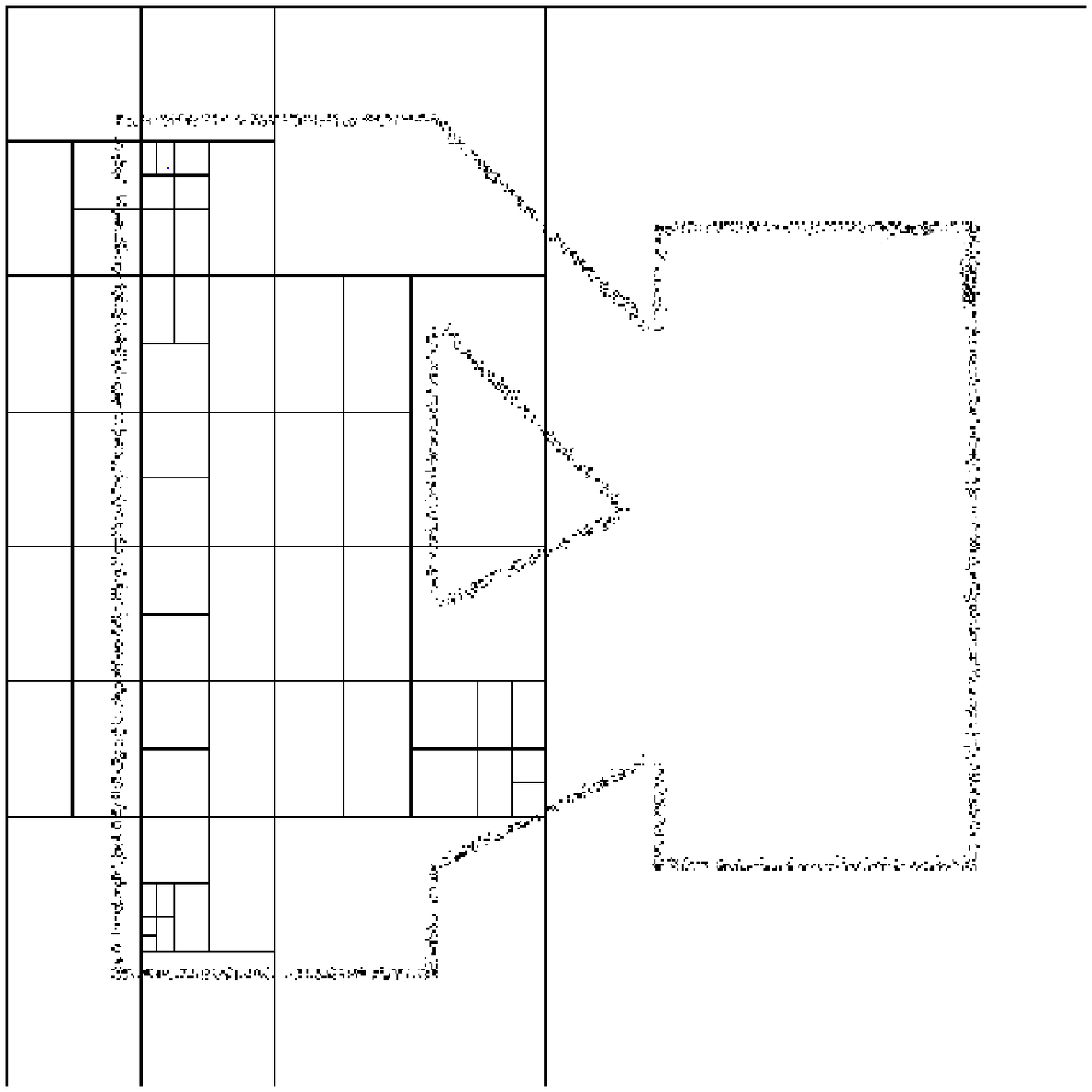}{7}
}
%\caption{Final State Aggregation of Module 1}
%\end{figure}
%\begin{figure}[htbp]
\centerline{
\image{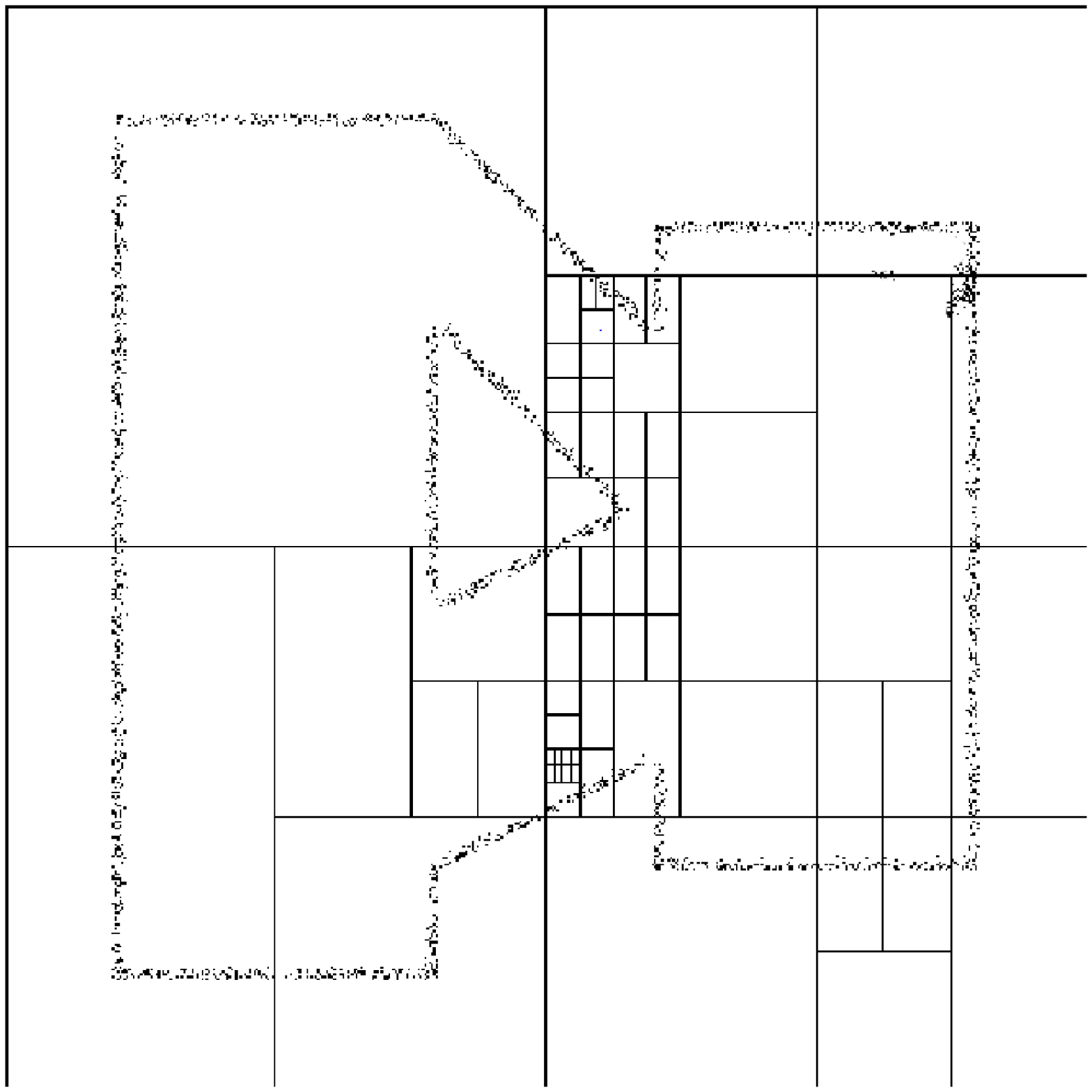}{7}
}
%\label{m1}
%\caption{Final State Aggregation of Module 2}
%\end{figure}
%\begin{figure}[htbp]
\centerline{\image{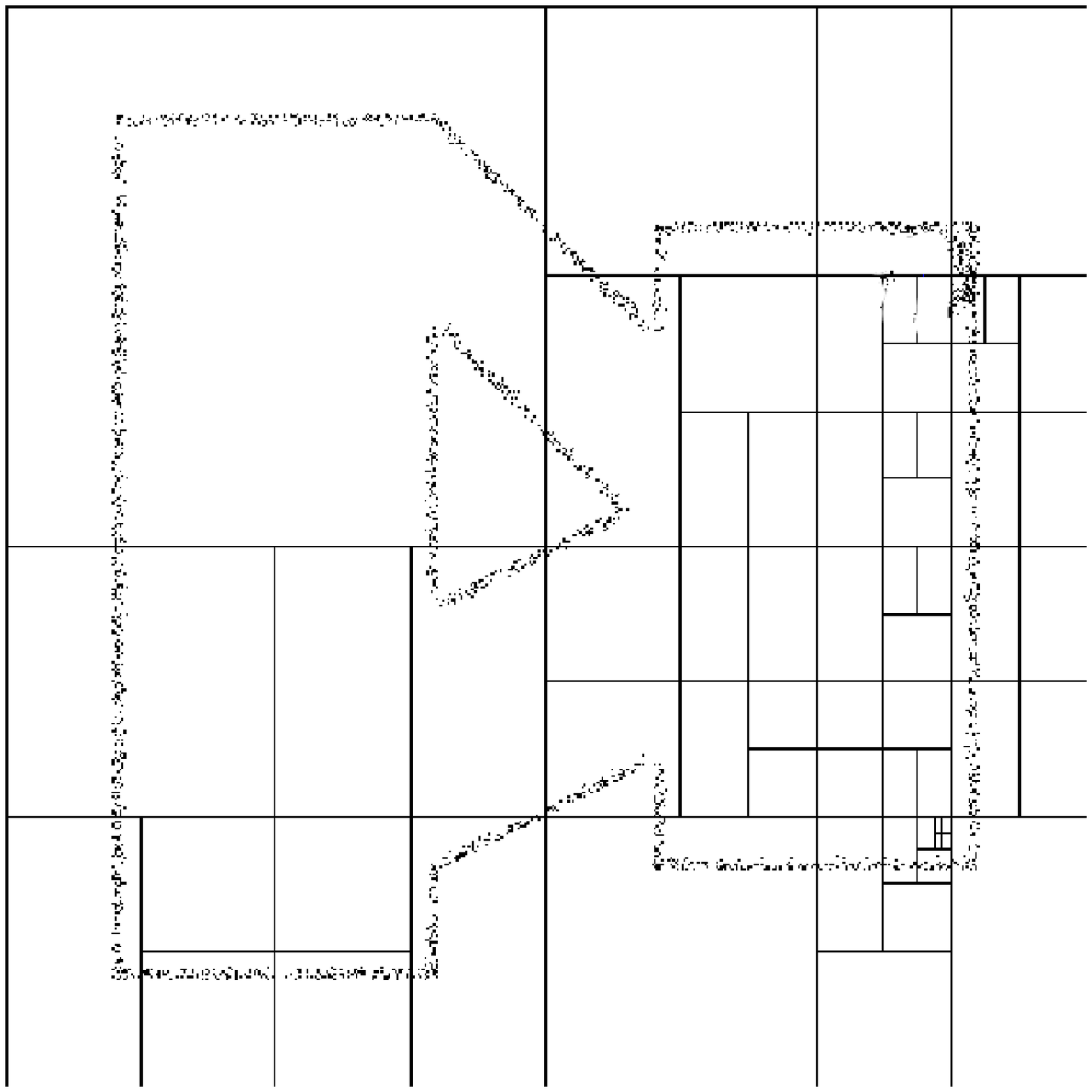}{7}}
%\label{m2}
\caption{Final State Aggregations of the three modules}
\label{m0}
\end{figure}
Finally, we see in figure \ref{m0} the state aggregations of
the resulting $3$ modules: we observe that, a module tends to describe
precisely the goal zones of its two automatically associated tasks.

\section{Conclusion}

In this paper, we have reviewed some recent results for sound
approximation in large state space Markov Decision Processes and
showed how they could be applied to the state aggregation scheme. We
have then showed how such results could be extended to an interesting
problem: The Modular Self-organizing of an autonomous agent. We have
formalized the problem of modular self-organization as a clustering
problem in the space of MDPs. We solved it using an on-line version
of the Dynamic Cluster algorithm. Finally, we have experimented this
approach in a continuous navigation framework, where a $3$-module agent
has to address $6$ tasks.

In future works, we will try to extend this general approach to more
powerful approximations schemes than the state aggregation approach
(which suffers from the curse of dimensionality). Furthermore, we
will investigate possible use in reinforcement learning, where the
parameters of an MDP have to be obtained by experience.

\bibliographystyle{plain}
\bibliography{./biblio}

\end{document}